\documentclass[times,twocolumn,review]{elsarticle}

\usepackage{medima}
\usepackage{framed,multirow}

\usepackage{latexsym}

\usepackage{url}
\usepackage{xcolor}
\usepackage{amsmath,amssymb,amsfonts}
\usepackage{hyperref}

\definecolor{newcolor}{rgb}{.8,.349,.1}
\newcommand{\tabincell}[2]{\begin{tabular}{@{}#1@{}}#2\end{tabular}}

\journal{Medical Image Analysis}

\begin{document}

\verso{Xiaoyu Bai \textit{et~al.}}

\begin{frontmatter}

\title{UAE: Universal Anatomical Embedding on Multi-modality Medical Images}%

\author[1]{Xiaoyu \snm{Bai}}

\author[4]{ Fan \snm{Bai}}
\author[2]{Xiaofei \snm{Huo}}
\author[3]{Jia  \snm{Ge}}
\author[5]{Jingjing \snm{Lu}}
\author[3]{ Xianghua  \snm{Ye}}
\author{Ke  \snm{Yan}}
\author[1]{Yong  \snm{Xia}\corref{cor1}}
\cortext[cor1]{Corresponding author.}
\ead{yxia@nwpu.edu.cn}

\address[1]{National Engineering Laboratory for Integrated Aero-Space-Ground-Ocean Big Data Application Technology, School of Computer Science and Engineering, Northwestern Polytechnical University, Xi’an 710072, China}
\address[2]{Beijing United Family Hospital, China}
\address[3]{The First Affiliated Hospital, Zhejiang University, Hangzhou, China}
\address[4]{Department of Electronic Engineering, The Chinese University of Hong Kong, Hong Kong,China}
\address[5]{Peking Union Medical College Hospital, Beijing, China}


\begin{abstract}
Identifying specific anatomical structures (\textit{e.g.}, lesions or landmarks) in medical images plays a fundamental role in medical image analysis. Exemplar-based landmark detection methods are receiving increasing attention since they can detect arbitrary anatomical points in inference while do not need landmark annotations in training. They use self-supervised learning to acquire a discriminative embedding for each voxel within the image. These approaches can identify corresponding landmarks through nearest neighbor matching and has demonstrated promising results across various tasks.
However, current methods still face challenges in: (1) differentiating voxels with similar appearance but different semantic meanings (\textit{e.g.}, two adjacent structures without clear borders); (2) matching voxels with similar semantics but markedly different appearance (\textit{e.g.}, the same vessel before and after contrast injection); and (3) cross-modality matching (\textit{e.g.}, CT-MRI landmark-based registration). To overcome these challenges, we propose universal anatomical embedding (UAE), which is a unified framework designed to learn appearance, semantic, and cross-modality anatomical embeddings. Specifically, UAE incorporates three key innovations: (1) semantic embedding learning with prototypical contrastive loss; (2) a fixed-point-based matching strategy; and (3) an iterative approach for cross-modality embedding learning. We thoroughly evaluated UAE across intra- and inter-modality tasks, including one-shot landmark detection, lesion tracking on longitudinal CT scans, and CT-MRI affine/rigid registration with varying field of view. Our results suggest that UAE outperforms state-of-the-art methods, offering a robust and versatile approach for landmark based medical image analysis tasks. Code and trained models are available at: \href{https://shorturl.at/bgsB3}{https://shorturl.at/bgsB3}.
\end{abstract}


\end{frontmatter}



\section{Introduction}
Identifying key anatomical structures (e.g. certain lesions or landmarks) from medical images plays a crucial role in the clinical workflow, as it helps with diagnosis, treatment planning, and other related procedures \cite{jiang2022cephalformer,quan2022images,zhong2019attention}. 
For instance, doctors need compare the lesions between baseline and follow-up CT scans of the same subject to identify changes \cite{yan2020learning,cai2021deep}. Besides, in an image registration pipeline, two images need to be roughly aligned before deformable warping, which can be achieved by matching a set of landmarks on both images \cite{balakrishnan2019voxelmorph,yin2022one}. 
Currently, most landmark detection methods rely on supervised learning \cite{o2018attaining,bier2018x,zhu2022datr}, which requires manually annotated key points as supervision. 
Despite the high cost of manual annotation, these methods are limited to handle predefined landmarks and not able to detect arbitrary structures such as lesions.

\begin{figure}[tb]
    \centering
    \includegraphics[scale=0.45]{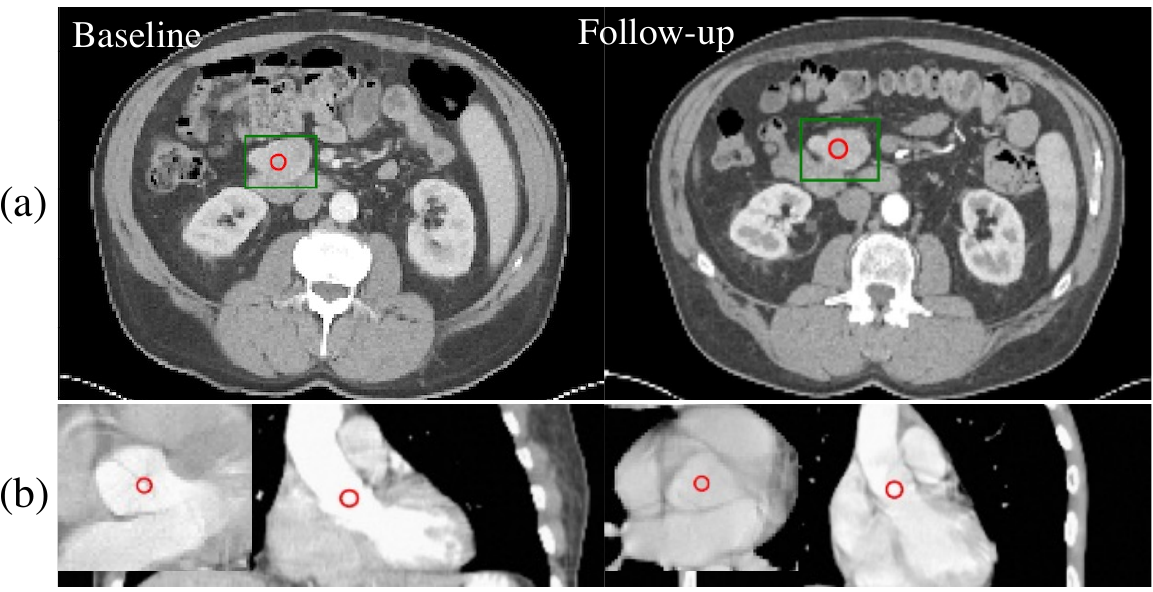}
    \caption{Two examples of exemplar-based landmark detection. (a) In intra-patient case, an lesion annotation from one CT is used to locate the corresponding lesion on the follow-up scan. (b) In inter-patient case, an anatomical structure annotation (aortic valve) from one CT is used to identify the same anatomical structure of another subject.}
    \label{fig:fig1}
\end{figure}

Recently, a variety of methods have emerged for landmark detection using only a single exemplar annotation \cite{yan2022sam,yao2021one,yao2022relative}. In these methods, landmark detection is formulated as a template-query matching problem, which aims to directly associate landmarks with exemplars based on their similarity in an anatomical embedding space. To achieve this, voxel-wise self-supervised learning pipelines are designed. Two augmented views of the same image are fed into a Siamese network to produce a dense feature map for each of them. The objective is to ensure that corresponding voxels in these two views render similar features, while non-corresponding voxels exhibit distinctive features. Given the inherent structural similarities in the human body across subjects, these methods are expected to produce similar features for the same anatomical structures among different subjects. Consequently, the annotations from one subject can be leveraged as exemplars to locate the corresponding structures in other subjects. 

\begin{figure}[tb]
    \centering
    \includegraphics[scale=0.50]{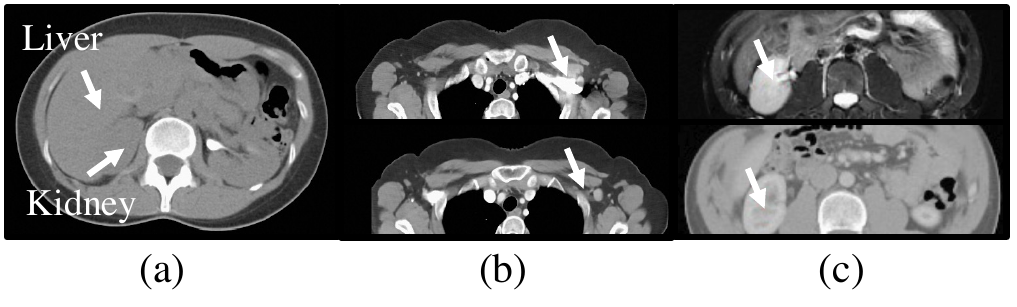}
    \caption{Hard cases for self-supervised anatomical embeddings. (a) In a non-contrast CT, the liver and kidney exhibit similar texture and intensity distribution, making it difficult to distinguish them using self-supervised appearance features. (b) The use of contrast agents greatly alters the appearance of vessels, leading to confusion with ribs.
    (c) Matching anatomical structures across modalities (e.g. MRI and CT).}
    \label{fig:fig2}
\end{figure}

One representative method within this context is the Self-supervised Anatomical eMbedding (SAM) \cite{yan2022sam}, which can produce a unique embedding for each voxel in a medical image such as a CT or MRI image. During inference, the nearest-neighbor (NN) matching technique is employed to locate the desired landmark based on the exemplar annotation. This approach has demonstrated promising results across various challenging tasks, including lesion tracking in longitudinal CT scans \cite{cai2021deep}, universal landmark matching (Fig.~\ref{fig:fig1}), and CT image registration\cite{liu2021same,li2023samconvex}. Despite its success, it still has two major limitations. 
First, the discriminatory ability of SAM stems from self-supervised similarity measures of contextual appearance. Hence it is difficult for SAM to distinguish the structures that share similar appearances but have different semantics (Fig.~\ref{fig:fig2} (a)) and to match the structures with similar semantics but distinct appearances (Fig.~\ref{fig:fig2} (b)). 
In such cases, SAM may produce significant matching errors.
Second, SAM is unable to perform multi-modality anatomy matching, which limits its utility in multi-modality applications such as multi-modality lesion tracking and registration (Fig.~\ref{fig:fig2} (c)).

To tackle these challenges, we propose universal anatomical embedding (UAE), a unified framework for learning appearance, semantic, and cross-modality embeddings. UAE introduces three key novelties.
\noindent \textbf{(1) Semantic embedding learning by prototypical supervised contrastive (SupCon) loss}: To address the problem of similar appearance with difference semantics, we incorporate a semantic head to generate a semantic embedding vector for each voxel. The semantic head is trained using the prototypical SupCon loss, which is specifically designed to facilitate supervised contrastive learning at the voxel level. Notably, we train the semantic head using publicly available organ segmentation datasets, making it easily accessible and adaptable. 

\noindent \textbf{(2) Fixed-point-based matching}: To address the problem of similar semantics showing different appearance, we propose a fixed-point-based iterative matching technique to replace the naive NN matching strategy. We leverage the relation between the target structure and its surrounding ``stable'' structures to produce more reliable matching. 

\noindent (3) We propose a novel iterative  \textbf{cross-modality embedding learning} method, which is designed to learn the correspondences between unregistered multi-modality images of the same subject without any manual annotation, even when they exhibit large field-of-view (FOV) differences.

We evaluate UAE on both single-modality and cross-modality tasks. The former includes tracking lesions in arbitrary organs of longitudinal CT scans \cite{cai2021deep} and landmark detection on CT images using a single exemplar \cite{yan2022sam}. The latter is cross-modality affine/rigid registration, with a specific focus on the scenarios with large FOV disparities among different modalities.
On single-modality tasks, UAE not only achieves a significant performance improvement over SAM (from 81.87\% to 84.06\%), but also outperforms state-of-the-art lesion tracking and exemplar-based landmark detection methods.
On the cross-modality registration task, UAE achieves significantly smaller mean distance errors than current methods (UAE: $5.1\pm 2.2$mm, C2FViT \cite{mok2022affine}: $13.4\pm11.6$mm), showing remarkable robustness under the conditions of significant FOV disparity. When applied to the downstream task of multi-modal tumor segmentation, our automatic method outperforms human-assisted traditional registration approach. UAE has also been integrated as the registration tool in our solution to the MICCAI Liver Lesion Diagnosis Challenge (LLD-MMRI 2023)\footnote{https://github.com/LMMMEng/LLD-MMRI2023}, where  we secured the 2nd position.
We have released the entire training and inference codes, together with easy-to-use demos and trained models to facilitate the community to build their applications with this useful anatomical embedding and matching tool.
\section{Related work}

\subsection{Exemplar-based medical landmark detection}
In exemplar-based (\textit{i.e.} one-shot) landmark detection methods, landmark detection in medical images is formulated as a template-query matching problem, with a central reliance on defining metrics to assess the similarity between voxels in template and query images. 
Most methods involve contrastive learning, which has a proven ability to generate similar features for visually similar images \cite{chen2020simple, caron2020unsupervised}.
For instance, Yao et al. \cite{yao2021one,yao2022relative} designed the Cascade Comparing to Detect (CC2D) method to generate similar embeddings at each location on an image and its augmentation and achieved promising results on X-ray datasets.
Based on CC2D, Quan et al. \cite{quan2022images} proposed the Sample Choosing Policy (SCP) method to select the most representative template image from a dataset.
Chen et al. \cite{chen2023unsupervised} developed the Local Discrimination (LD) method, which learns contrastive self-supervised dense embeddings and generates segmentation masks through clustering simultaneously. They achieved promising performance across several 2D medical image datasets. 

SAM \cite{yan2022sam} extends exemplar-based medical landmark detection to 3D images like CT and MRI.  It is designed to capture anatomical information at the voxel level and generate similar embeddings for corresponding body parts, thereby facilitating template-query matching. SAM employs a coarse-to-fine contrastive learning process, where the coarse level learns global appearances, and the fine level learns local textures. Given a 3D image, SAM extracts two partially-overlapping patches and augments them through scaling and intensity transform.  Voxels that appear in the same location on both patches are considered positive pairs, while other voxels (beyond a threshold to the positive pairs) on both patches are treated as negative samples. SAM deliberately selects hard and diverse negative samples to form the negative pairs. Finally, the InfoNCE loss \cite{oord2018representation} is applied to reduce the distance between positive pairs and push negative pairs apart. This results in all voxels on the 3D image  having a distinct embedding representation and the same location on different augmented views having similar embeddings. Due to the high structural similarity of the human body across individuals, SAM can also output similar embeddings for the same anatomical structure on different subject's scans. A recent study by Vizitiu et al.\cite{vizitiu2023multi} showed that SAM-style method can also benefit from the optional supervision, such as automatically extracted anatomical landmarks.

Essentially, these methods optimize the same objective, which is defined based on treating the same location on two augmented views as a positive pair and considering two different locations as a negative pair. Therefore, they share the weakness of SAM, whereas our approach is designed to address these weaknesses.

\subsection{Multi-modality data alignment}
\begin{figure}[tp]
    \centering
    \includegraphics[scale=0.32]{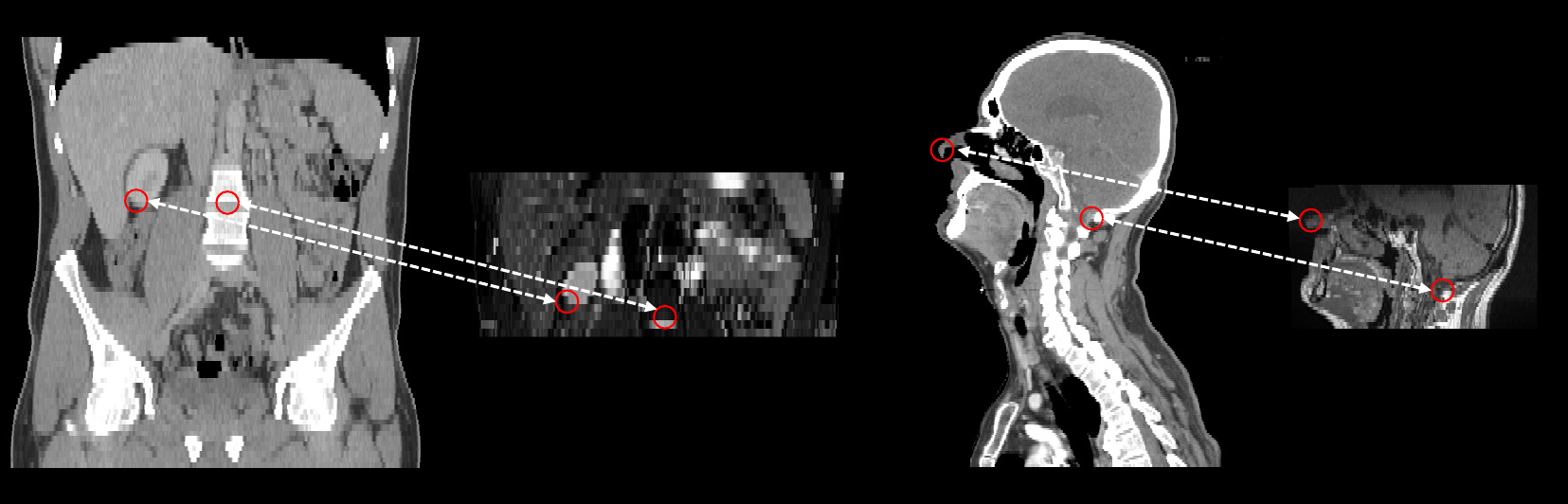}
    \caption{Two examples of multi-modality medical images with substantially different FOVs. Left: Abdominal CT and MRI images; Right: Head and neck CT and MRI images. We can achieve the roughly alignment of different images by matching the same anatomical landmarks on them.
    }
    \label{fig:fov}
\end{figure}

The diagnostic accuracy can be significantly enhanced by utilizing the complementary information from aligned multi-modality medical images. A notable example is that the aligned MRI and CT images can empower radiotherapists for precise treatment planning \cite{khoo2000comparison}. Typically, multi-modality medical images are aligned in two steps: a global affine or rigid transformation \cite{mok2022affine,huang2021coarse,hoffmann2021synthmorph} and a deformable registration \cite{balakrishnan2019voxelmorph,mok2020large}.
However, in our experiments, existing affine/rigid registration methods cannot work well when two images have substantially different FOVs (Fig.~\ref{fig:fov} and ~\ref{fig:regis}). 
In such circumstances,  align the correspond body part and crop them into a similar FOV is needed.

This alignment issue can be addressed by exemplar-based landmark detection methods, which include matching sampled points on fixed and moving images followed by automatic image cropping. For single-modality images, SAME-affine\cite{liu2021same,li2023samconvex} leveraged SAM \cite{yan2022sam} to match corresponding points for affine registration, demonstrating impressive robustness in handling FOV differences. For multi-modality images, however, training a SAM-style model with cross-modality capabilities to drive SAME-affine may encounter a chicken-and-egg issue \cite{yin2022one}, since, on one hand, we require registered multi-modality images to train SAM and learn the correspondence and, on the other hand, we need the corresponding SAM landmarks for accurate image registration with a large FOV disparity.

\begin{figure*}[tp]
    \centering
    \includegraphics[scale=0.42]{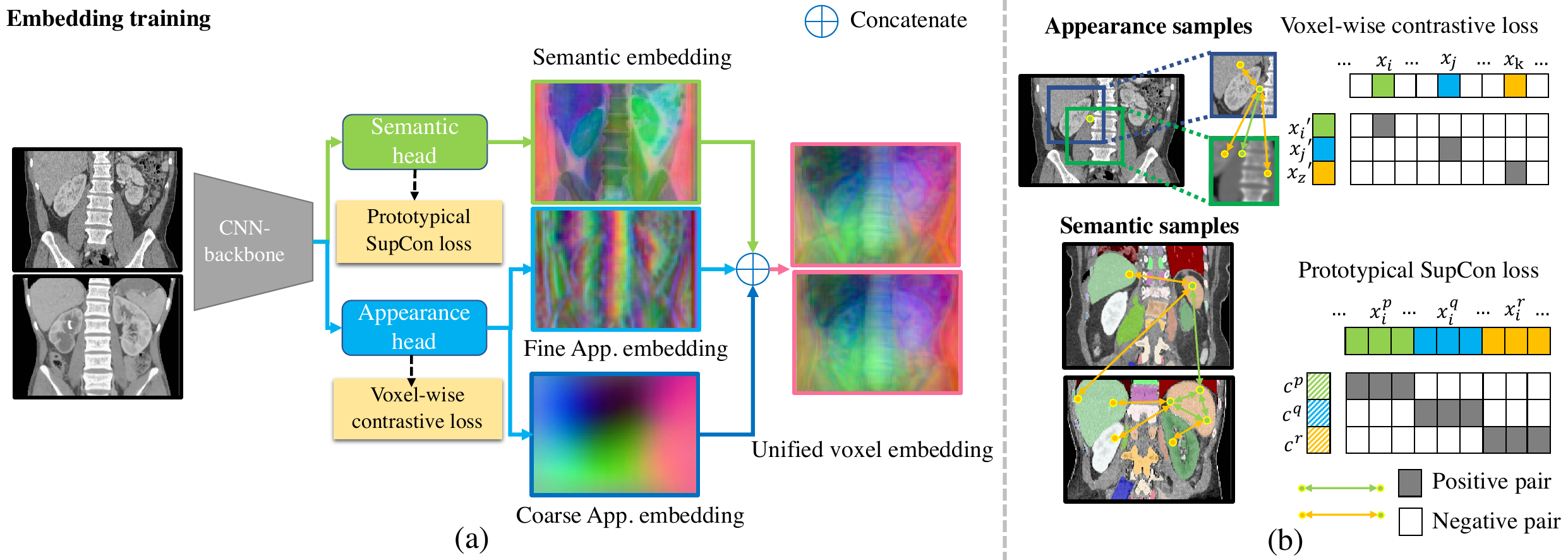}
    \caption{Overview of SEA. (a) SEA contains a semantic head and an appearance head. The semantic head generates semantic embeddings, the appearance head is responsible for generating both coarse and fine appearance (App.) embeddings.  The dimension of embeddings is reduced to to 3 using PCA, and each embedding can be shown as a RGB image. (b) Illustration of voxel-wise contrastive loss and prototypical supervised contrastive loss, where $x_i$ and $x_i'$ are the embeddings of the same voxel on two augmented patches, and $c^p$ and $x_i^p$ represent the prototype and instance embeddings of class $p$.}
    \label{fig:framework-a}
\end{figure*}

Our solution to this chicken-and-egg issue involves a two-phase learning process. Initially, we train a modality-agnostic UAE  by applying strong contrast augmentation on single modality images. This strategy is used to simulate the appearance difference in multi-modality images (\textit{e.g.}, CT and MRI). Subsequently, we use this modality-agnostic UAE as a seed to initiate an iterative process that alternates between cross-modality embedding learning and cross-modality registration. In the end, we obtain both a cross-modality UAE model and registered images. For new data, we can directly employ the trained cross-modality UAE for affine/rigid registration. The resulting images can then be fed into available deformable registration methods for voxel-wise alignment. By combining these two components, we create a robust, fully automated registration tool.

\section{Method}
\begin{figure*}[htp]
    \centering
    \includegraphics[scale=0.48]{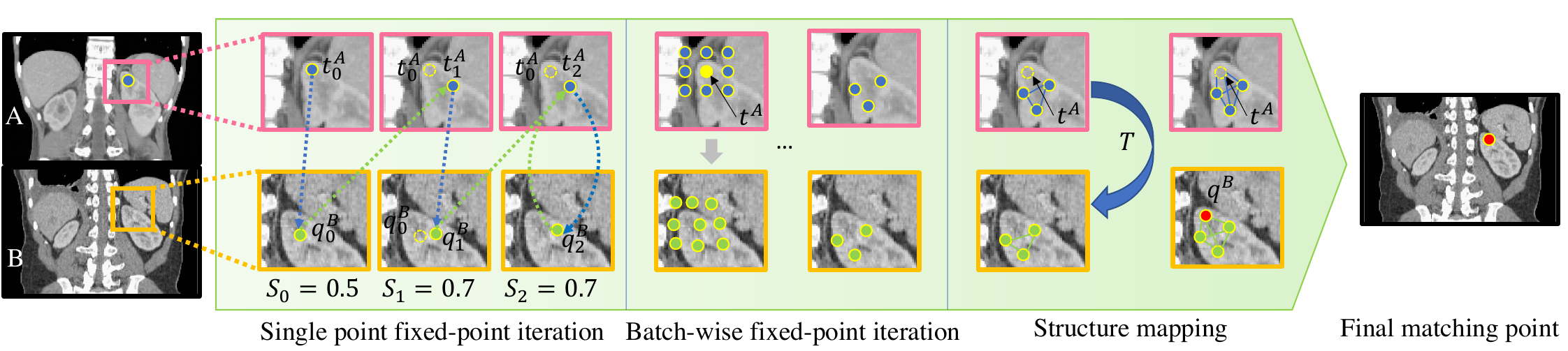}
    \caption{Illustration of fixed-point-based matching method. We first show single point fixed-point iteration. Starting from the template point $t^A=t_0^A$, we identify a fixed point $t_2^A$ with a high similarity score using iterative NN matching. The process is then extended to the grid points around $t^A$ using batch-wise fixed-point iteration to obtain all fixed points (different starting points may converge to the same fixed point). Subsequently, we learn how the fixed points structure around $t^A$ is mapped to the corresponding structure in $B$ through an affine transform $T$. By considering $t^A$ as an element of the fixed points structure in $A$, we can compute the matching point as $q^B=Tt^A$, which is more accurate and reliable than the initial NN matching result $q_0^B$.
    }
    \label{fig:framework-b}
\end{figure*}

The proposed UAE consists of UAE-S and UAE-M, which are designed for single- and multi-modality embedding learning, respectively. UAE-S can be further divided into two components: Semantic-Enhanced Anatomical embedding model (SEA) and fixed-point-based matching. The former is designed for appearance and semantic embedding learning, and the latter is a more reliable alternative for nearest neighbor (NN) matching. The illustrations of SEA, fixed-point-based matching, and UAE-M are displayed in Fig. \ref{fig:framework-a}, Fig. \ref{fig:framework-b}, and Fig. \ref{fig:crosssam}, respectively. We now delve into the details of each module.

\subsection{SEA}
SEA includes a semantic branch and an appearance branch, which share the same convolutional neural network (CNN) backbone
(see Fig. \ref{fig:framework-a} (a)). 
Note that we also tried to use transformer backbones but observed no significant performance gain.
The appearance branch employs the same training strategy as the original SAM, where two overlapped and randomly augmented patches  are fed into the CNN-backbone followed by the appearance head to generate the appearance embedding for each voxel. Let the embeddings of the positive pair be denoted by $\{\boldsymbol{x}_i, \boldsymbol{x}_i'\}$. The appearance branch aims to minimize the following voxel-wise contrastive loss
\begin{equation}
\small
\mathcal{L}_{\text{app}}=-\sum_{i=1}^{n_{\text{pos}}} \log \frac{\exp \left(\boldsymbol{x}_i \cdot \boldsymbol{x}_i' / \tau_{\text{a}}\right)}{\exp \left(\boldsymbol{x}_i \cdot \boldsymbol{x}_i' / \tau_{\text{a}}\right)+\sum_{j=1}^{n_{\text{neg}}} \exp \left(\boldsymbol{x}_i \cdot \boldsymbol{x}_j / \tau_{\text{a}}\right)},
\end{equation}
where $n_{\text{pos}}$ and $n_{\text{neg}}$ denote the number of positive and negative pairs, respectively, and $\tau_{\text{a}}$ is the temperature parameter.

The self-supervised appearance head is able to learn the difference between two distinct body structures based on their appearances. However, it cannot distinguish challenging cases such as adjacent tissues and organs, which share similar intensity distribution and texture (see Fig.~\ref{fig:fig2}(a)). To overcome this issue, we need higher-level semantic supervision to differentiate different tissues and organs. 

We utilize a semantic branch to produce a fixed-length semantic embedding and treat it as a supplement to the appearance embedding. Our goal is to make the embeddings of the same organ closer than those from different organs. We can leverage any public organ segmentation dataset with arbitrary organ annotations. However, directly using the voxel-level supervised contrastive (SupCon) loss \cite{khosla2020supervised} is very expensive, since its complexity is $O(n^2)$, where $n$ is the number of voxels in a patch ($\sim$300K). To address this issue, we design a prototypical SupCon loss by replacing the voxel-voxel pairs with prototype-voxel pairs. During training, given the predicted embeddings produced by the semantic head $\{\boldsymbol{x}_i^p\},i\in[1,n_p],p\in[1,K]$, where $\boldsymbol{x}_i^p$ represents $i$th voxel embedding with semantic label $p$, $n_p$ is the number of voxels with label $p$, $K$ is the number of semantic classes (i.e. organ labels), we formulate the prototypical SupCon loss as

\begin{equation}
\small
\mathcal{L}_{\text{s}}=\sum_{p=1}^{K}-\dfrac{1}{n_p}\sum_{i=1}^{n_p}\log \dfrac{\exp{( \boldsymbol{c}_p \cdot  \boldsymbol{x}_i^p/ \tau_{\text{s}})}}{\sum_{q=1}^K\sum_{j=1}^{n_q}\exp{(\boldsymbol{c}_p \cdot \boldsymbol{x}_j^q/ \tau_{\text{s}})}},
\end{equation}
where $\boldsymbol{c}_p=\frac{1}{n_p} \sum_{i=1}^{n_p}\boldsymbol{x}_i^p$ is the prototype of class $p$, and $\tau_{\text{s}}$ is the temperature parameter. In contrast to the original SupCon loss, prototypical SupCon loss reduces the complexity from $O(n^2)$ to $O(nK)$, enabling its usage on dense prediction tasks.
We apply the $\mathcal{L}$-2 normalization to the embeddings produced by both appearance and semantic heads, so that they can be directly concatenated and used as a unified embedding vector.

\subsection{Fixed-point-based matching}
After learning the SEA model, we also need a robust and accurate point matching strategy. Current exemplar-based methods compute the inner product of template and query embeddings and uses NN matching. However, when the target structure in the query image is missing or significantly altered (see Fig.~\ref{fig:fig2}(b)), this simple strategy may not be accurate. 

Therefore, we propose an iterative structural inference method (see Fig.~\ref{fig:framework-b}) to improve matching performance, which takes into account the consistency of a match and the relationship between the target landmark and its surrounding structures. Feeding two images $A$ and $B$ into SEA, we obtain the voxel embeddings $\boldsymbol{X}^A = \{\boldsymbol{x}_i^A\}$ and $\boldsymbol{X}^B = \{\boldsymbol{x}_i^B \}$, where $i$ is the index of pixels. Let the template embedding vector on $A$ be denoted by $\boldsymbol{x}_t^A$. We can find the corresponding query embedding on $B$ through NN matching
\begin{equation}
\boldsymbol{x}_q^B = \mathrm{argmax}_{i\in B}(\boldsymbol{x}_t^A \cdot \boldsymbol{x}_i^B).
\end{equation}
For simplicity, we ignore the notation of embedding $\boldsymbol{x}$ and represent the template point and its NN matching point on~$B$ as: $t^A=t_0^A$ and $q_0^B$. We have established the correspondence $t_0^A \rightarrow q_0^B $. 
Let us consider the reverse process. Starting from $q_0^B$, can the NN matching method give us $t_0^A$? If yes, we can conclude that we have a consistent forward-backward matching and that is probably a good matching. However, if the reverse process maps to another point, \textit{i.e.}, $q_0^B \rightarrow t_1^A$ and $t_1^A \neq t_0^A$, then the first NN match is probably not reliable, since the similarity score of $S(q_0^B \rightarrow t_1^A)$ is larger than $S(q_0^B \rightarrow t_0^A)$.

Formally, the forward-backward process can be formulated as
\begin{equation}
t_{i+1}^A=f(t_i^A,\boldsymbol{X}^A,\boldsymbol{X}^B)\triangleq t_i^A \rightarrow q_i^B,q_i^B \rightarrow t_{i+1}^A.
\end{equation}
Mathematically, a fixed point of a function is an element that is mapped to itself by the function\cite{granas2003fixed}. Therefore, a forward-backward consistent matching identifies a fixed point of $f$ since $t_0^A = t_1^A = f(t_0^A)$. For the matchings where $t_1^A \neq t_0^A$, although it is not a fixed point, we can always use it as a starting point to find a fixed point using the fixed-point iteration. Specifically, for any $t_0^A$, we compute a sequence of its $f$ mappings
\begin{equation}
{t_0^A, t_1^A = f(t_0^A), t_2^A = f(t_1^A), t_{i+1}^A = f(t_i^A), ...}.
\end{equation}
This process gradually increases the similarity score until, after $n_\text{{fix}}$ iterations, the sequence converges to $t_{i+1}^A = t_i^A$ for all $i \geq n_\text{{fix}}$. In light of the fixed-point iteration, we propose an approach to locate the matching of $t^A$ by considering all fixed points surrounding it. Our method begins with selecting an $L^3$ cubic region centered at the template point $t^A$ and applying batched fixed-point iteration. Subsequently, we preserve fixed points with offsets to $t^A$ below a threshold $\tau_{dis}$. We treat these fixed points on image $A$ as a structured element, and view their corresponding points on image $B$ as another structure. We then estimate an affine transform $T$ to describe their mapping relation. By including $t^A$ in the fixed-point structure on $A$, we can compute the matching point as $q^B=Tt^A$.
When a query point has a drastically different appearance, its NN matching result becomes unreliable. Our fixed-point based matching method can adaptively find highly reliable points surrounding the query point and aggregate this structural information to give the final matching result.

\subsection{UAE-M}
\begin{figure*}[tp]
    \centering
    \includegraphics[scale=0.48]{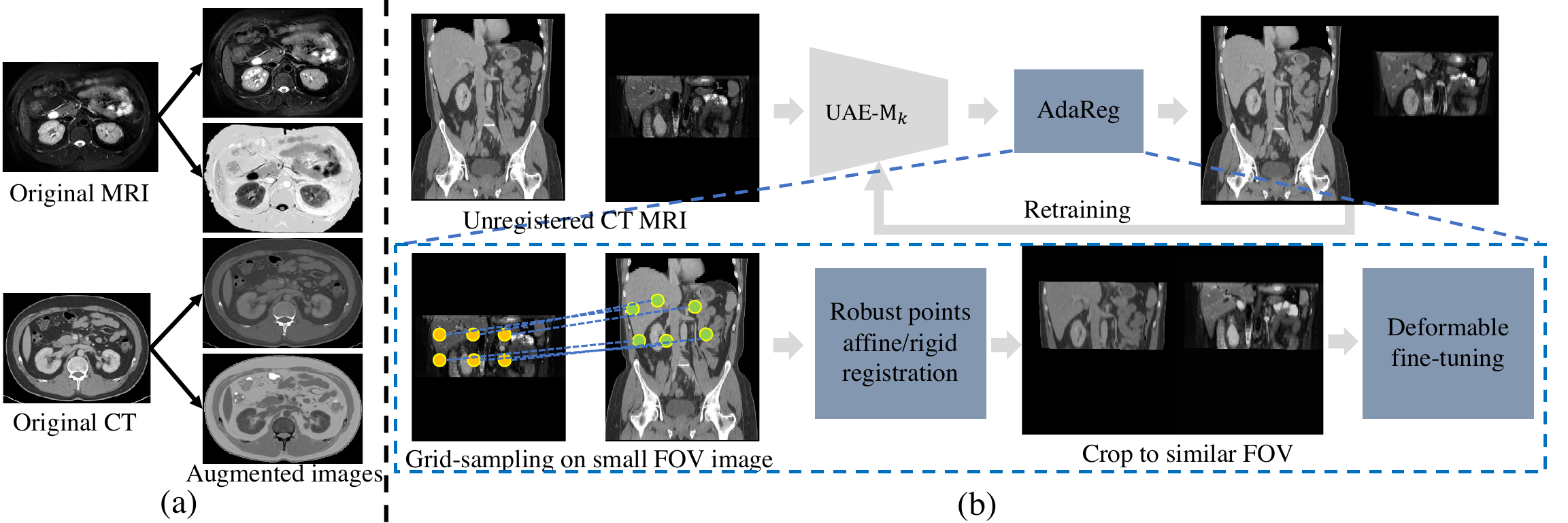}
    \caption{Illustration of UAE-M Framework. (a) shows the results of using aggressive intensity augmentation on MRI and CT images. (b) The iterative training procedure.}
    \label{fig:crosssam}
\end{figure*}
In multi-modality scenarios, for example, we have a dataset comprising both CT and MRI images of each subject. Notably, each CT and MRI pair is not registered and could exhibit significant FOV differences, which is common in raw clinical data (CT scans often have larger FOV than MRI, see Fig.~\ref{fig:fov}). We aim to learn multi-modality anatomical embeddings on this dataset, and thus find the correspondence between CT and MRI images. A straightforward solution is to register the CT and MRI images of each subject and then learn the cross-modality correspondence. Existing registration methods, however, can hardly deal with the image pair with significant FOV differences.

To address this issue, we introduce an iterative method for learning the UAE-M model, which shares the same design as the SEA model but employs a distinct learning approach. To maintain simplicity, we utilize only the appearance head of SEA in UAE-M, while the semantic head can be seamlessly integrated if necessary. Drawing inspiration from the concept of modality-agnostic learning \cite{billot2023synthseg}, we initially train an UAE-M model that is moderately modality-independent by applying strong, or even aggressive, contrast augmentation.

Specifically, we randomly apply non-linear intensity transformation \cite{zhou2021models} and intensity reversal to each image. This process generates visually unrealistic intensity images (as shown in Fig.~\ref{fig:crosssam}(a)), but it preserves the anatomical structures. Furthermore, we include the random affine transformation, resizing, and blurring in our augmentation pipeline. During training, we first pick one CT or MRI image randomly, crop two overlapping patches, then apply different aggressive augmentations to these two patches, and finally use them to train the initial UAE-M (UAE-M$_0$) model. In this way, we force the model to learn the features that are independent of intensity and focus on higher-level structural similarity. As both CT and MRI are structural images, this UAE-M$_0$ can roughly match the regions with strong structural information.

Nonetheless, directly using this aggressively augmented UAE-M$_0$ for single-point matching is still locally inaccurate. Instead, we 
can estimate a reliable global affine/rigid transform from the matchings of multiple points. In SAME-affine \cite{liu2021same}, evenly spaced points were employed to calculate the affine matrix for two images. Since our CT-MRI pair is from the same subject, we favor computing the rigid transform matrix to prevent unwanted stretching.  Our task is thus to solve a rigid fitting problem of two 3-D point sets. Given two 3-D point sets ${p_i}$ and ${p_i'}$, $i=1,\cdots ,N$, we assume 
\begin{equation}
p_i^{\prime}=R p_i+t+n_i,
\end{equation}
where $R$ is a rotation matrix, $t$ is a translation vector, and $n_i$ is instance noise. We aim to find $R$ and $t$ that minimize 
\begin{equation}
    \Sigma^2=\sum_{i=1}^N\left\|p_i^{\prime}-\left(R p_i+t\right)\right\|^2.
\end{equation}
We utilize a quick and robust non-iterative method \cite{arun1987least} to compute the rigid transfer matrix, which enables us to map the scan with a small FOV to its corresponding scan with a large FOV and crop the latter. Due to the potential inaccuracy of the rigid transfer, we dilate the body region of the small FOV scan to avoid the risk of cropping overlapping regions from two scans. 
After cropping the scan, both scans have similar FOVs. Then we apply a widely-used deformable registration method called DEEDS \cite{heinrich2013mrf} to refine the results. This process is referred to as AdaReg (see Fig.~\ref{fig:crosssam} (b)).

After registering CT-MRI pairs using AdaReg, we can learn the cross-modality correspondence. 
For the fine-level learning, we select
$N_{\text{pos}}^f$ positive pairs from the registered regions with overlapping areas and choose $N_{\text{neg}}^f$ points to act as negative samples. Additionally, we choose $N_{\text{fov}}^f$ points from the non-overlapping area of the large FOV scan. Then the loss function is defined as
\begin{equation}
\small
\mathcal{L}=-\sum_{i=1}^{N_{\text{pos}}^f} \log \frac{\exp \left(\boldsymbol{x}_i \cdot \boldsymbol{x}_i' / \tau_{c}\right)}{\exp \left(\boldsymbol{x}_i \cdot \boldsymbol{x}_i' / \tau_{c}\right)+\sum_{j=1}^{N_{\text{neg}}^f+N_{\text{fov}}^f} \exp \left(\boldsymbol{x}_i \cdot \boldsymbol{x}_j /\tau_{c}\right)},
\end{equation}
where $x_i$ and $x_i'$ are the embeddings of a positive pair, $x_j$ represents the negative sample, and $\tau_{c}$ is a temperature parameter. Similarly, for the coarse-level learning, we select $N_{\text{pos}}^c$ positive pairs and $N_{\text{neg}}^c$ negative samples for each positive pair. To fully utilize the data, we also include augmented intra-modality data as input and using the self-supervised training method. 
However, as the alignment of CT-MRI pairs is based on the aggressive augmented UAE-M$_0$, erroneous correspondences may exist in imperfectly aligned cases. To improve the accuracy, we iterate the following two steps: run AdaReg using  UAE-M$_{k-1}$ embedding and learn UAE-M$_k$ with image pairs from AdaReg, where $k$ is the iteration number. In each iteration, we reduce the margin of the dilated body region of the small FOV image, resulting in a closer FOV for the cropped pairs and a more accurate deformable fine-tuning process. Our final UAE-M is obtained when this process converges after several iterations.

\section{Experiments}
\subsection{Datasets}

We trained UAE-S on two public datasets.
The NIH-Lymph Node (NIH-LN) dataset\cite{yan2022sam} includes 176 chest-abdomen-pelvis CT scans, and the Total Segmentator dataset \cite{wasserthal2022totalsegmentator} contains 1204 CT images with the labels of 104 anatomical structures.
We evaluated UAE-S on two tasks: lesion tracking and landmark detection using CT. 
The lesion tracking task aims to match the same lesion on longitudinal CT scans. We used the publicly available deep longitudinal study (DLS) dataset \cite{cai2021deep}, which contains 3008, 403, and 480 lesion pairs for training, validation, and testing, respectively. 
For landmark detection, we used the ChestCT dataset, which contains the  contrast-enhanced (CE) and non-contrast (NC) CT scans of 94 subjects.

For UAE-M, we use two in-house datasets for training and test. The head-and-neck (HaN) dataset consists of 120 paired T1 MRI and non-contrast CT images, which were not co-registered. Each MRI image has a voxel size of around 0.5$\times$0.5$\times$6 $\text{mm}^3$, and each CT image has a voxel size of around 1$\times$1$\times$3 $\text{mm}^3$. The MRI images have a limited FOV and mainly capture the region between the nose and the second cervical vertebra, while the CT images include the regions from the top of head to a portion of the lungs. We used 100 cases for training and reserved 20 for test. 
The abdomen (ABD) dataset contains 98 pairs of T2 MRI and non-contrast CT scans. The voxel sizes of CT and MRI images are 0.7$\times$0.7$\times$5 mm and 0.8$\times$0.8$\times$8 mm, respectively. The MRI images have a small FOV that covers only a portion of liver and kidney, while the CT images encompass regions from the bottom of lung to the thighbone. We used 80 cases for training and 18 cases for test.

\subsection{Performance Metrics}
The accuracy of lesion tracking was assessed using the Center Point Matching (CPM) method \cite{tang2022transformer,cai2020deep}. A match is deemed correct if the Euclidean distance between the predicted and ground truth centers is smaller than a threshold, which is set to either 10mm or the lesion radius\cite{tang2022transformer,cai2020deep}. Other performance metrics are the Mean Euclidean Distance (MED) between the predicted and ground truth centers and its projections in each direction (referred to as MEDX, MEDY, and MEDZ).
For the ChestCT landmark matching, we use the same setting as SAM by calculating the mean distance of 19 predefined landmarks using template-query matching.

We assessed the performance of cross-modality affine/rigid registration by comparing the voxel-level MED of the same landmark on the registered pairs. Specifically, we annotated 12 landmarks on both CT and MRI images on the HaN dataset, including the lacrimal gland, the endpoint of temporomandibular joint, the top and bottom of C2 spine, the middle point of jawbone, and the intersection of the lateral pterygoid muscle and upper jawbone. On the ABD dataset, we annotated 6 landmarks, including the top and bottom points of liver and spleen, as well as the top points of kidneys. 

\begin{table*}[htp]
\centering
\caption{Comparison of lesion tracking methods and our UAE-S on test set of DLS dataset.} \label{tab1}
\resizebox{\textwidth}{!}{
\begin{tabular}{l|c|c|c|c|c|c}
\hline
Method & \tabincell{c}{CPM$@$10$mm$ \\ (DLT- / TLT-Results)} & \tabincell{c}{CPM$@$Radius \\ (DLT- / TLT-Results)} \ & \tabincell{c}{MED$_{X}$ \\ ($mm$)} \ & \tabincell{c}{MED$_{Y}$ \\ ($mm$)} \ & \tabincell{c}{MED$_{Z}$ \\ ($mm$)} \ & \tabincell{c}{MED \\ ($mm$)} \ \\
\hline
SiamRPN++, 2019~\cite{li2019evolution} & $68.85/-$ & $80.31/-$ & $3.8\pm4.8$ & $3.8\pm4.8$ & $4.8\pm7.5$ & $8.3\pm9.2$ \\
LENS-LesaNet, 2020~\cite{yan2020learning,yan2019holistic} & $70.00/-$ & $84.58/-$ & $2.7\pm4.8$ & $2.6\pm4.7$ & $5.7\pm8.6$ & $7.8\pm10.3$ \\
DLT-Mix, 2021~\cite{cai2021deep} & $78.65/-$ & $88.75/-$ & $3.1\pm4.4$ & $3.1\pm4.5$ & $4.2\pm7.6$ & $7.1\pm9.2$ \\
DLT, 2021~\cite{cai2021deep} & $78.85/-$ & $86.88/-$ & $3.5\pm5.6$ & $2.9\pm4.9$ & $4.0\pm6.1$ & $7.0\pm8.9$ \\
TLT, 2022 \cite{tang2022transformer} & $-/87.37$ & $-/95.32$ & $3.0\pm6.2$ & $3.7\pm5.2$ & $\mathbf{1.7\pm2.1}$ & $6.0\pm7.7$ \\
\hline
Affine, 2016~\cite{marstal2016simpleelastix} & $48.33/-$ & $65.21/-$ & $4.1\pm5.0$ & $5.4\pm5.6$ & $7.1\pm8.3$ & $11.2\pm9.9$ \\
VoxelMorph, 2018~\cite{balakrishnan2018unsupervised} & $49.90/-$ & $65.59/-$ & $4.6\pm6.7$ & $5.2\pm7.9$ & $6.6\pm6.2$ & $10.9\pm10.9$ \\
DEEDS, 2013~\cite{heinrich2013mrf} & $71.88/-$ & $85.52/-$ & $2.8\pm3.7$ & $3.1\pm4.1$ & $5.0\pm6.8$ & $7.4\pm8.1$ \\
SAM, 2022 \cite{yan2022sam} & $81.87/86.14$ & $91.56/95.41$ & $2.6\pm3.8$ & $2.3\pm2.9$ & $4.0\pm5.6$ & $6.2\pm6.9$ \\
Vizitiu et al., 2023~\cite{vizitiu2023multi}&$83.13/-$&$91.87/-$&$2.9\pm6.0$&$2.2\pm3.2$&$3.1\pm3.9$&$5.9\pm7.1$\\
UAE-S (proposed) & $\mathbf{84.06/89.27}$ & $\mathbf{93.02/96.77}$ & $\mathbf{2.3\pm3.1}$ & $\mathbf{2.1\pm2.7}$ & $3.6\pm4.7$ & $\mathbf{5.4\pm5.7}$ \\
\hline
\end{tabular}
}
\begin{minipage}{18cm}
\small The methods in the upper part use lesion annotations in training, and thus have task-specific supervision. The methods in the lower part do not use lesion annotations. We observed discrepancies between the evaluation code presented in the TLT paper \cite{tang2022transformer} and the DLT \cite{cai2021deep}. To ensure fairness, we report the results on both evaluation codes using the format of DLT- / TLT-Results.
\end{minipage}
\end{table*}

\begin{figure}[htp]
    \centering
    \includegraphics[scale=0.44]{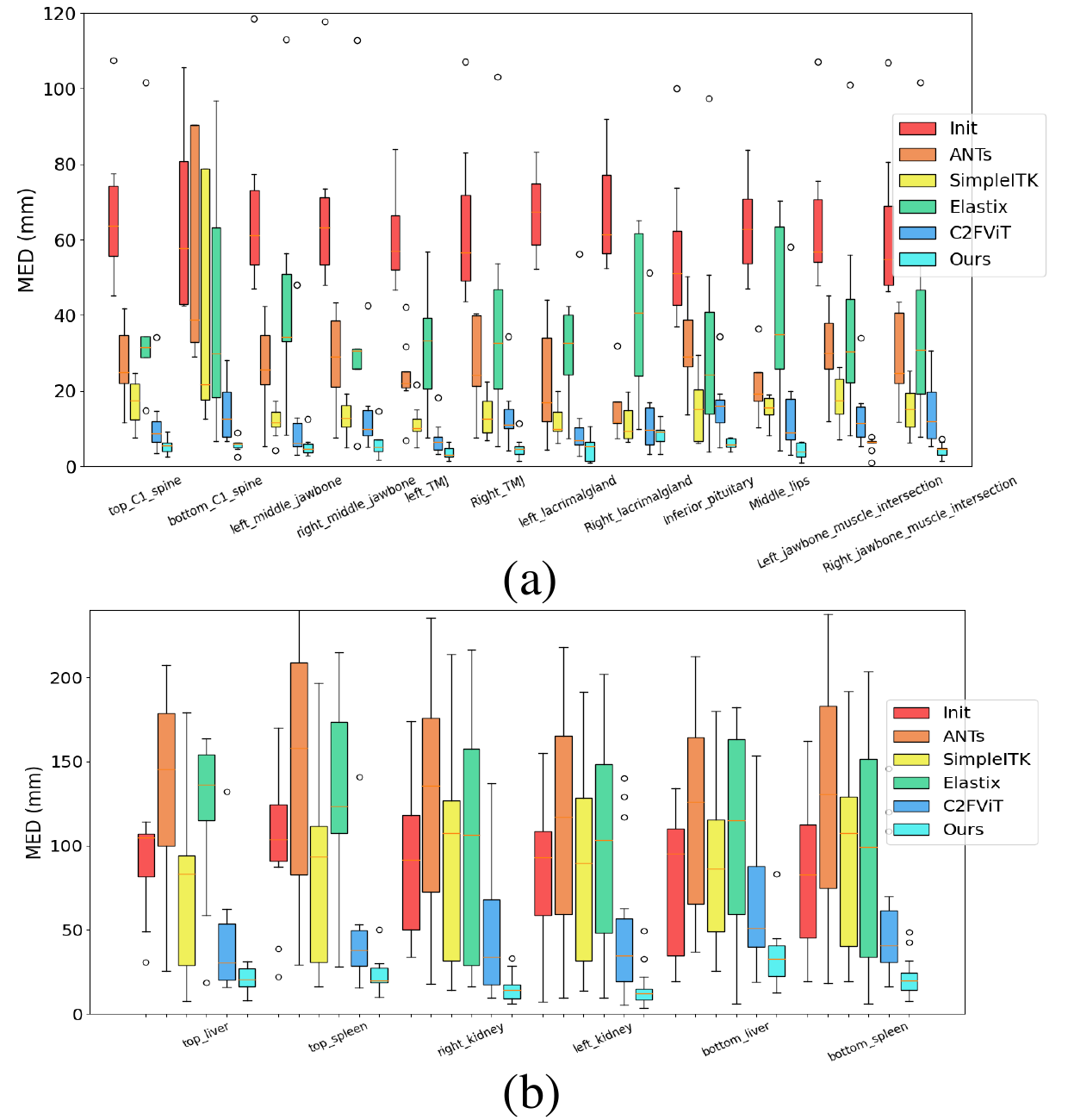}
    \caption{Box-plot of the MED error of CT-MRI rigid registration on (a) HaN and (b) ABD datasets.}
    \label{fig:boxplot-all}
\end{figure}

\subsection{Implementation Details}
The proposed UAE was implemented using PyTorch v1.9 and MMDetction v1.20. We adopted 3D ResNet18 as our CNN-backbone and 3D feature pyramid network (FPN) as our semantic and appearance head. The embedding length of each head is set as 128. For the appearance head, we generated the coarse level embedding and fine level embedding. For the semantic head, we only generated the fine-level embedding. To save GPU memory, the size of output is the half of input volume size, we then used trilinear interpolation to re-slice the output to the original size.
The network was optimized by SGD with a momentum of 0.9. The learning rate was set to 0.02, the batch size was set to 5, and the temperature $\tau_{\text{a}}$ and $\tau_{s}$ were set to 0.5. All CT volumes have been re-sliced to the isotropic resolution of 2$\times$2$\times$2 $\text{mm}^3$. For fixed-points matching, we set $L=5$. We used random rotation, scaling (in $[0.8,1.2]$), noise, and blurring for data augmentation. During training of UAE-M, we iterated in each mini-batch with self-supervised learning and registered pair learning. For the self-supervised part, we cropped patches to a size of 96 x 96 x 32 voxels and used a batch size of 4. For registered pairs, the input is the entire overlapped region along with a randomly selected region in the non-overlapped area. Due to the limited GPU memory, we only put one pair of data in each mini-batch. For fine-level embedding learning, we selected $N_{\text{pos}}^f=200$ positive pairs, and, for each positive pair, we randomly selected $N_{\text{neg}}^f=500$ and $N_{\text{fov}}^f=100$ samples from the non-overlapped region.
For coarse-level embedding learning, we set $N_{\text{pos}}^c=100$, $N_{\text{neg}}^c=200$, and $\tau_{\text{c}}=0.5$. For AdaReg, we set the dilation margin of the body mask as 10, 5, and 1 for each iteration.
\begin{table*}[htb]
\center
\caption{Comparison of anatomical landmark detection methods and our UAE-S on ChestCT dataset. Same to original SAM \cite{yan2022sam} paper, we reported mean error $\pm$ std. and max error.} \label{tab2}
\begin{tabular}{l|c|c|c|c}
\hline
Method & \tabincell{c}{CE-CE} & \tabincell{c}{NC-NC} \ & \tabincell{c}{CE-NC} \ & \tabincell{c}{NC-CE} \\
\hline
 \multicolumn{5}{c}{On 19 test cases \cite{yan2022sam} } \\
\hline
Affine~\cite{marstal2016simpleelastix} ~~~~~ & 8.4$\pm$5.2~32.9 & 8.5$\pm$5.3~33.1&-&-\\\hline
DEEDS~\cite{heinrich2013mrf} &4.6$\pm$3.3~18.8& 4.7$\pm$3.4~24.4&-&-\\\hline
VoxelMorph~\cite{balakrishnan2018unsupervised}  &7.3$\pm$3.6~20.1&7.4$\pm$3.7~20.2&-&-\\\hline
SAM \cite{yan2022sam} &4.3$\pm$3.0~16.4&4.5$\pm3.0$~18.5&-&-\\
\hline
 \multicolumn{5}{c}{On all 94 cases} \\
\hline
SAM \cite{yan2022sam} & 4.8$\pm$3.4~23.6&4.6$\pm$3.5~25.9&4.7$\pm$3.5~44.0&5.2$\pm$3.9 29.6\\\hline
UAE-S&\textbf{4.0$\pm$2.5~16.1}&\textbf{3.9$\pm$2.4~17.0}&\textbf{4.0$\pm$2.5~17.2}&\textbf{4.0$\pm$2.5~16.9}\\\hline
\end{tabular}
\end{table*}
\begin{figure*}[htp]
    \centering
    \includegraphics[scale=0.33]{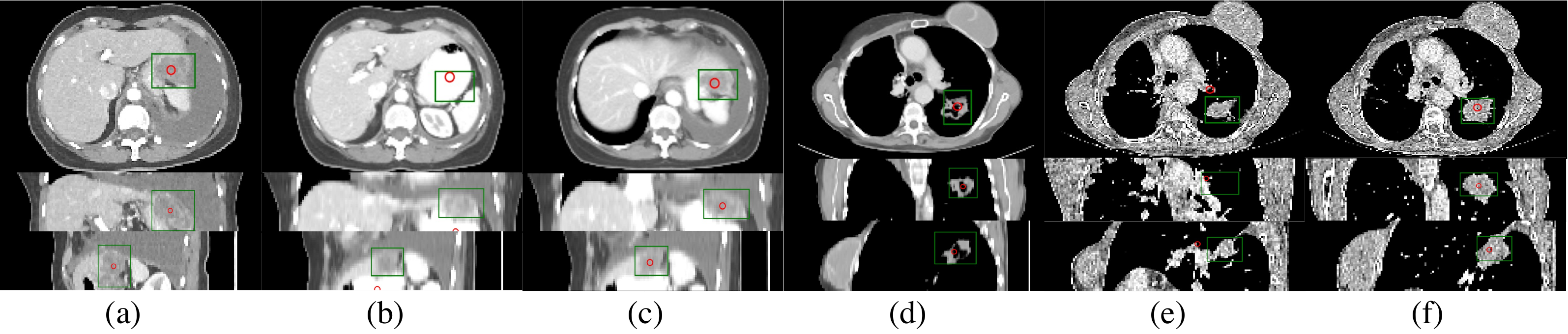}
    \caption{A visual comparison of lesion tracking results of SAM and UAE-S. We show two challenging cases with deformed organs and lesions, and altered contrast and image quality. Green boxes show the true lesion regions. Red circles indicate the true and tracked lesion centers. (a) and (d) are template (baseline) CT images. (b)(e) and (c)(f) are tracking results of SAM and UAE-S on the follow-up CT scans, respectively.}
    \label{fig:dlt}
\end{figure*}
\begin{table*}[htb]
\centering
\caption{Ablation study of UAE-S on full DLS dataset \cite{cai2021deep} using evaluation code of TLT \cite{tang2022transformer}.} \label{tab3}
\centering
\begin{tabular}{l|c|c|c|c|c|c}
\hline
Method & \tabincell{c}{CPM$@$ \\ 10$mm$} & \tabincell{c}{CPM$@$ \\ Radius} \ & \tabincell{c}{MED$_{X}$ \\ ($mm$)} \ & \tabincell{c}{MED$_{Y}$ \\ ($mm$)} \ & \tabincell{c}{MED$_{Z}$ \\ ($mm$)} \ & \tabincell{c}{MED \\ ($mm$)} \ \\
\hline
SAM \cite{yan2022sam} & $88.94$ & $93.44$ & $2.4\pm 3.0$ & $2.5\pm3.1$ & $3.6\pm3.8$ & $5.8\pm5.0$ \\
UAE-S w/o Fix-point-based matching & $89.37$ & $93.99$ & $2.3\pm2.6$ & $2.5\pm2.8$ & $3.5\pm3.7$ & $5.6\pm4.5$ \\
UAE-S w/o Semantic head & $89.66$ & $94.44$ & $2.2\pm2.4$ & $2.3\pm2.8$ & $3.4\pm3.7$ & $5.3\pm4.4$ \\
UAE-S & $\mathbf{91.05}$ & $\mathbf{95.45}$ & $\mathbf{2.2\pm2.2}$ & $\mathbf{2.1\pm2.2}$ & $\mathbf{3.2\pm3.4}$ & $\mathbf{5.1\pm3.8}$ \\
\hline
\end{tabular}

\end{table*}

\begin{table}[htp]
\small
\caption{Accuracy of nasopharyngeal carcinoma gross tumor volume (GTV) segmentation.} \label{tab:NPC}
\begin{tabular}{c|c|c|c|c}
\hline
Method  &Dice &HD95  & Precision & Recall   \\
\hline
CT-only &74.1$\pm$8.5 &8.8$\pm$5.7
&76.4$\pm$16.3
&\textbf{77.6$\pm$15.6}

\\\hline
CT-MRI-M &75.0$\pm$8.2 &8.5$\pm$4.8&79.2$\pm$15.4&75.9$\pm$14.6
\\\hline
CT-MRI-C&\textbf{76.4$\pm$8.1
} &\textbf{7.7$\pm$3.6
}&\textbf{83.0$\pm$13.8
}&74.7$\pm$14.0
\\
\hline
\end{tabular}
\begin{minipage}{9cm}
\small CT-only means using only manually cropped CT to do segmentation. CT-MRI-M means manually cropping the CT volume for the rough region between the nose and the second cervical vertebra, followed by registering MRI to CT volume. CT-MRI-C denotes our UAE-M method, which involves automatic cropping of the region based on the FOV of MRI volume. For all methods, we used DEEDS for deformable registration and nnUNet \cite{Isensee2021nnUNet} for segmentation.
\end{minipage}
\end{table}
\subsection{Results}
Table \ref{tab1} displays the lesion tracking results on the DLS test set. We compared with supervised lesion tracking methods \cite{li2019evolution,yan2020learning,yan2019holistic,cai2021deep,tang2022transformer}, registration methods \cite{marstal2016simpleelastix,balakrishnan2018unsupervised,heinrich2013mrf},  SAM \cite{yan2022sam} and the improved SAM \cite{vizitiu2023multi}. It reveals that UAE-S outperforms all competing methods, though it does not use any task-specific supervision. Note that the organ masks used to train the semantic head contain no lesion information. 
The lesion tracking results produced by SAM and UAE-S were visualized in Fig. \ref{fig:dlt}. It shows that the difficult cases for SAM can be effectively handled by UAE-S.

Table \ref{tab2} gives the performance of landmark detection on the ChestCT dataset. Previously in \cite{yan2022sam}, the performance was examined exclusively on 19 test cases, focusing solely on intra-phase settings.
By contrast, we conducted validation on all 94 cases, testing both intra- and inter-phase matching scenarios. It shows in Table \ref{tab2} that the proposed UAE-S outperforms SAM across all settings consistently.
\begin{figure*}[htp]
    \centering
    \includegraphics[scale=0.54]{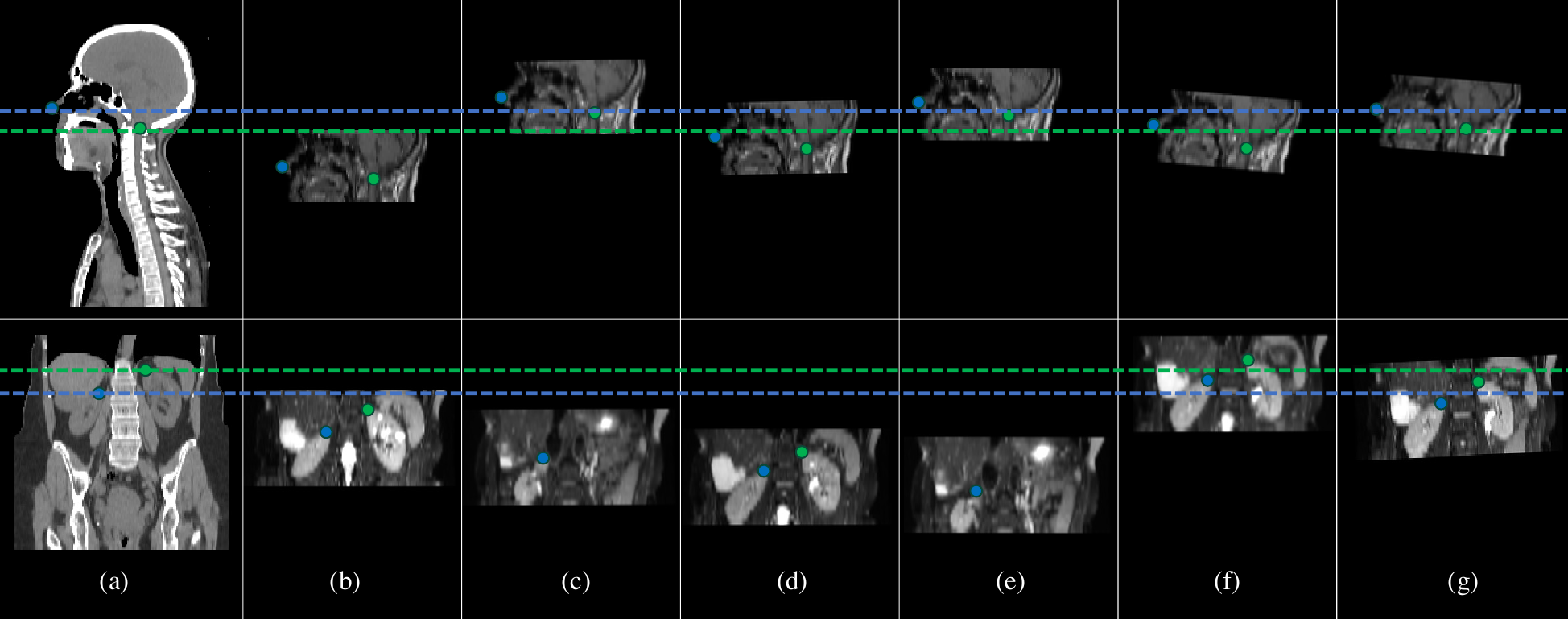}
    \caption{Visualization of different methods for rigid registration on HaN and ABD datasets. (a) Fixed images. (b) Moving images. (c) Results of ANTs. (d) Results of Elastix. (e) Results of SimpleITK. (f) Results of C2FViT. (g) Results of our UAE-M. For each image, we highlight the same two landmarks using blue and green dots. The green and blue dashed lines indicate the parallel projections of the landmarks on the fixed images. }
    \label{fig:regis}
\end{figure*}
\begin{figure*}[htp]
    \centering
    \includegraphics[scale=0.6]{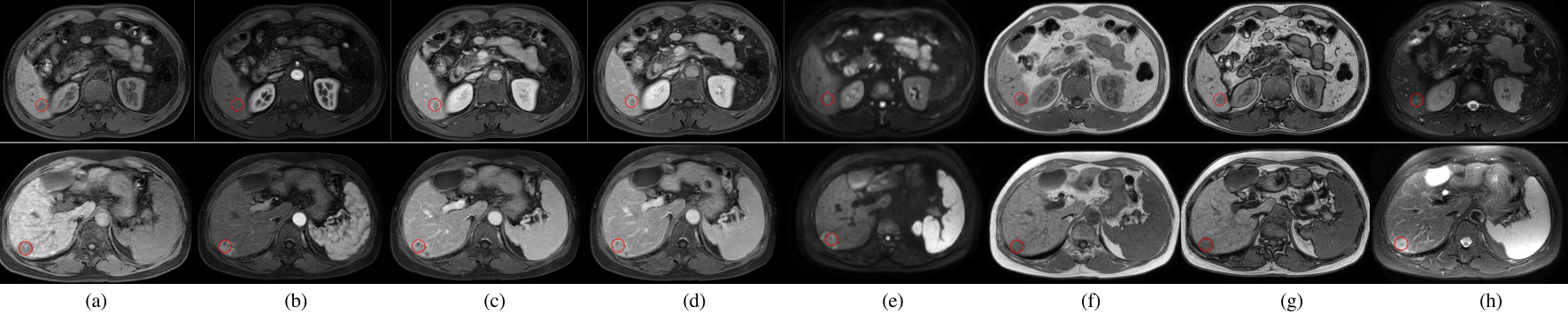}
    \caption{Two lesion matching examples using UAE-M on eight MRI modalities from the LLD-MMRI challenge. We present two examples.  (a) Template pre-contrast phase images. (b) Arterial phase images. (c) Venous phase images. (d) Delay phase images. (e) DWI images. (f) In phase images. (g) Out phase images. (g) T2WI images. The lesions are marked in red circles. }
    \label{fig:lld}
\end{figure*}

\begin{figure*}[htp]
    \centering
    \includegraphics[scale=0.43]{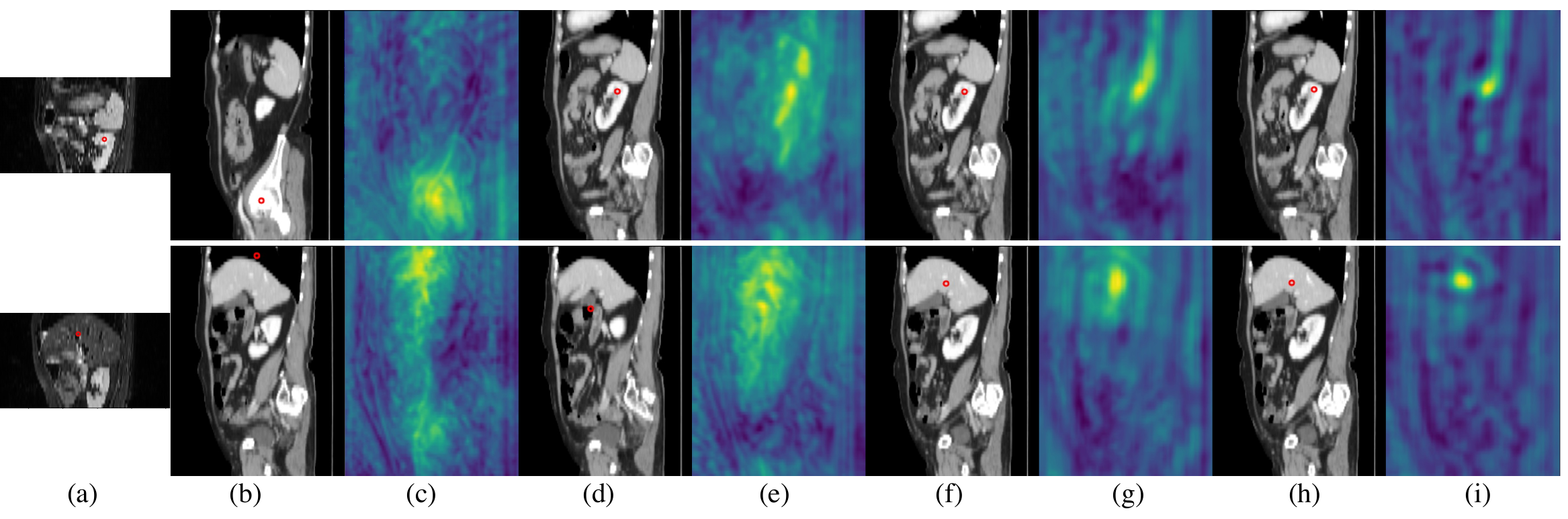}
    \caption{Template-query matching by different models. (a) Template image, the red dots reperesent the template points. (b)(c) The matching results and similarity map using SAM method. (d)(e) UAE-M$_0$ results. (f)(g) UAE-M$_1$ results. (h)(i) UAE-M$_2$ results.}
    \label{fig:csampoint}
\end{figure*}

\begin{figure*}[t]
    \centering
    \includegraphics[scale=0.85]{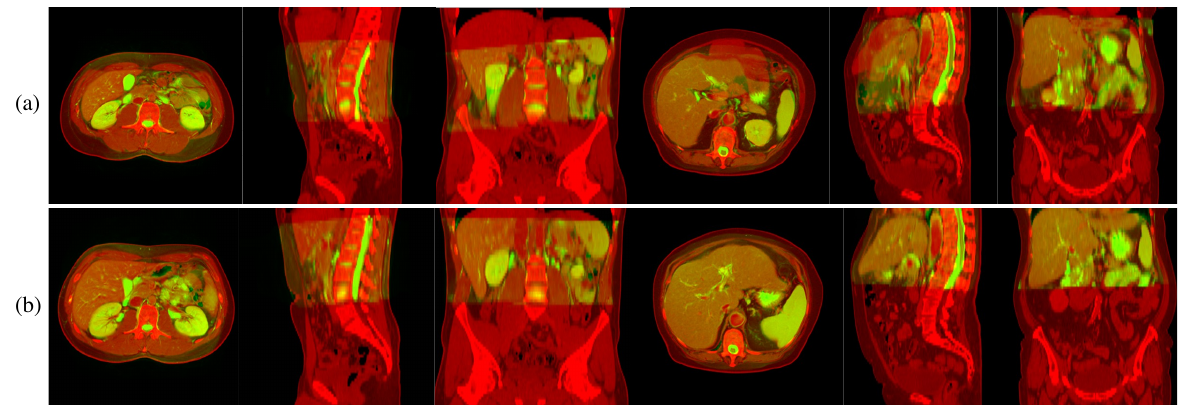}
    \caption{Examples of cross-modality registration under FOV differences. MRI images (green) were overlaid on CT images (red). (a) UAE-M$_0$+DEEDS. (b)UAE-M$_2$+DEEDS.
    }
    \label{fig:aggcross}
\end{figure*}

For the cross-modality registration task, we conducted a comprehensive comparison between our UAE-M and C2FViT \cite{mok2022affine} and three widely used conventional affine/rigid registration methods (\textit{i.e.}, ANTs \cite{avants2009advanced}, Elastix \cite{klein2009elastix} and SimpleITK \cite{lowekamp2013design}).
Among competing methods, C2FViT is a top-performing learning-based registration approach, in which a local normalized cross-correlation (NCC) loss is utilized for intra-modality registration. Since the NCC loss may not be ideal for cross-modality scenarios, we followed the suggestion of the authors and replaced it with the local mutual information loss \cite{maes2003medical,chen2022transmorph} in our implementation of C2FViT. 
Since each CT-MRI pair is from the same subject, rigid registration is a more suitable choice. The box-plot results of rigid registration were depicted in Fig.~\ref{fig:boxplot-all}. It shows that conventional methods struggle to handle the cases with significant FOV differences and hence yield poor performance in both HaN and ABD datasets. C2FViT stands out by reducing MED on both datasets. Our UAE-M excels even further, consistently reducing MED for all landmarks on both datasets. It indicates that our UAE-M is capable of providing a solid initial alignment for subsequent deformable registration in the highly challenging cross-modality and diverse-FOV scenario. The visualization of the registration results were shown in Fig.~\ref{fig:regis}.

UAE-M can be readily used in downstream tasks. We evaluated the performance of nasopharyngeal carcinoma (NPC) segmentation on the HaN dataset using UAE-M as the initial affine step. The results were presented in Table \ref{tab:NPC}. It show that our method outperforms manual cropping, enabling fully automated CT-MRI-based lesion segmentation. UAE-M has also been used to learn the correspondence on multi-phase MRI images. Fig. \ref{fig:lld} shows an example of liver lesion matching on multi-phase MRI images from the LLD-MMRI challenge dataset. We achieved second place in the challenge thanks to the reliable performance of UAE-M. 
\section{Discussion}
\subsection{Ablation Study}
Since we do not use DLS training set in model training, we conducted ablation study of UAE-S on the full DLS dataset. 
The results were presented in Table \ref{tab3}. We showed SAM as baseline and it yielded robust performance. UAE-S w/o Fix-point-based matching (SEA model + NN matching)  achieved CPM@Radius  93.99\%. UAE-S w/o Semantic-head  achieved a performance to 94.44\%. These results demonstrate that both contributions we made in this work are effective and outperform the original SAM model. Finally, by combining these two techniques, the proposed UAE-S further improved CPM@Radius to 95.45\%.

In Table \ref{tab:abd}, we presented the overall MED of 6 landmarks on the ABD test set after each iteration for UAE-M. The results reveal that the initial UAE-M$_{0}$ significantly reduces the overall MED from 87 to 36mm.  After two retraining iterations, the final UAE-M$_{2}$ reduces MED to 25.7mm. Adding deformable warping to DEEDS after the rigid registration step can further reduce MED. In Fig.~\ref{fig:aggcross}, we show the examples of UAE-M+DEEDS registration. 
These results suggest the effectiveness and robustness of the proposed method on cross-modality registration tasks.

\begin{table}[htp]
\small
\centering
\caption{MED (mm) after each UAE-M iteration on ABD dataset. Top row: UAE-M  rigid transform results. Bottom row: Adding DEEDS for deformable registration.} \label{tab:abd}
\begin{tabular}{c|c|c|c}
\hline
Initial& UAE-M$_{0}$  & UAE-M$_{1}$  & UAE-M$_{2}$ \\
\hline
$87.2\pm39.6$&$35.8\pm18.0$ &$29.5\pm14.9$& $25.7\pm11.6$\\
\hline
- &$17.8\pm17.2$&$11.8\pm 10.6$&$11.0\pm9.1$\\\hline
\end{tabular}
\end{table}

\subsection{Analysis of the Fixed-point-based Matching}
In the fixed-point-based matching approach, we select an $L^3$ cubic region centered at the template point as seeds to identify fixed-point pairs. Table \ref{tab:l} presents the lesion tracking results using different 
$L$ values. The case where $L=0$ corresponds to UAE-S without fixed-point-based matching. As shown in the table, even when selecting fixed-point pairs in a  small region ($L=3$), performance gains can be observed. Considering both performance and computational cost, $L=5$ is the optimal choice.

\begin{table}[htp]
\small
\centering
\caption{The performance comparison of different 
$L$ values in the fixed-point-based matching on the full DLS dataset.} \label{tab:l}
\begin{tabular}{c|c|c|c|c}
\hline
$L$ value &0 & 3  & 5  & 7 \\
\hline
CPM@10mm&89.37& 90.02  &91.05& 90.99\\
\hline
CPM@Radius&93.99&94.65 &95.45&  95.27\\\hline
\end{tabular}
\end{table}

\subsection{Comparison between SAM and UAE}

UAE-S and SAM target the same tasks while UAE-S demonstrates superior performance. On the other hand, UAE-M is designed to focus on the multi-modality matching task, which SAM and current exemplar-based methods cannot handle effectively. To illustrate this, we present the template-query matching results and corresponding similarity maps in Fig. \ref{fig:csampoint}. We show two examples: one involves a template point on the kidney, which has strong shape (structure) clues, and the other is in the liver, with less structural information. As shown in the figure, SAM fails on both examples. The similarity maps reveal that SAM cannot even highlight the corresponding region in the kidney example, although it appears to be easier than the liver example.

The UAE-M$_0$ model, trained using aggressive contrast augmentation, can correctly match the kidney example but fails in the liver example. Regarding the similarity maps, the highlighted regions are roughly correct but not concentrated, which causes the single-point template-query matching to be inaccurate. Therefore, we employ the multi-points-based AdaReg method to find better alignment and  to learn UAE-M$_k$ model iteratively. As evident in Fig. \ref{fig:csampoint} (g) and (i), the iterative training results in more concentrated highlights in the similarity maps, indicating more confident and accurate matching.
\section{Conclusion}

Self-supervised exemplar-based landmark detection is an emerging topic as it does not need landmark annotations to train the model and can detect arbitrary anatomical points accurately and conveniently. We propose UAE to address the limitations of current methods, improving the accuracy in intra-modality tasks and extending the ability to the multi-modality scenario. UAE introduces three key innovations: (1) Semantic embedding learning with a prototypical supervised contrastive loss to equip the anatomical embeddings with semantic information; (2) A fixed-point-based structural matching mechanism for more precise inference; and (3) A robust iterative pipeline for cross-modality anatomical embedding learning. UAE has shown promising results on various medical image tasks, including lesion tracking in longitudinal CT scans, one-shot landmark detection, and cross-modality rigid registration with downstream tasks. We look forward to more applications to be developed with our released codes and models.

\section*{Acknowledgments}
This work was supported in part by the National Natural Science Foundation of China under Grants 62171377, and in part by the National Key R\&D Program of China under Grant 2022YFC2009903 / 2022YFC2009900.
The authors would like to thank Tony C. W. Mok, Zi Li, Minfeng Xu, Le Lu, and Dakai Jin for their invaluable help and suggestions.

\bibliographystyle{model2-names.bst}\biboptions{authoryear}
\bibliography{refs}

\end{document}